\documentclass[10pt]{article}
\usepackage[a4paper, total={6in, 9in}]{geometry}

\usepackage[boxruled,linesnumbered,vlined,inoutnumbered]{algorithm2e}
\SetKwInOut{Parameter}{Parameters}

\usepackage{amsmath}
\usepackage{amssymb}
\usepackage{mathtools}
\usepackage{amsthm}
\usepackage{mathrsfs}
\usepackage{enumitem}
\usepackage[backend=bibtex8,
            sorting=none]{biblatex}
\usepackage{algorithm2e}

\usepackage{graphicx}
\usepackage{subfig}
\usepackage{booktabs}

\DeclareMathOperator*{\argmax}{arg\,max}
\DeclareMathOperator*{\pr}{\text{Pr}}
\DeclareMathOperator*{\ex}{\mathbb{E}}

\newcommand{\defeq}{\vcentcolon=}

\usepackage{xcolor}
\definecolor{light-grey}{rgb}{0.9,0.9,0.9}
\definecolor{dark-red}{rgb}{0.4,0.15,0.15}
\definecolor{dark-blue}{rgb}{0,0,0.7}

\usepackage{hyperref}
\hypersetup{
    colorlinks, linkcolor={dark-blue},
    citecolor={dark-blue}, urlcolor={dark-blue}
}

\title{Report on Learning Rules for Discrete-valued Networks}
\title{Unbiased Weight Maximization}
\author{Stephen Chung\thanks{University of Massachusetts Amherst, Department of Computer Science, \texttt{minghaychung@umass.edu}}}
\date{March 2022} 
\begin{document}
\maketitle 
\section{Overview} 

A biologically plausible method for training an Artificial Neural Network (ANN) involves treating each unit as a stochastic Reinforcement Learning (RL) agent, thereby considering the network as a team of agents. Consequently, all units can learn via REINFORCE, a local learning rule modulated by a global reward signal, which aligns more closely with biologically observed forms of synaptic plasticity \cite[Chapter~15]{sutton2018reinforcement}. Nevertheless, this learning method is often slow and scales poorly with network size due to inefficient structural credit assignment, since a single reward signal is broadcast to all units without considering individual contributions.

Weight Maximization \cite{chung2022learning}, a proposed solution, replaces a unit's reward signal with the norm of its outgoing weight, thereby allowing each hidden unit to maximize the norm of the outgoing weight instead of the global reward signal. In this research report, we analyze the theoretical properties of Weight Maximization and propose a variant, Unbiased Weight Maximization. This new approach provides an unbiased learning rule that increases learning speed and improves asymptotic performance. Notably, to our knowledge, this is the first learning rule for a network of Bernoulli-logistic units that is unbiased and scales well with the number of network's units in terms of learning speed.

\section{A single Bernoulli-logistic unit}

We first focus on the learning rules for a single Bernoulli-logistic unit with no inputs\footnote{To generalize the learning rules to a unit with input, we can multiply the update of the bias by the input to compute the update of the weight associated with that input.}: denoting the activation value of the unit by $H$, sigmoid function as $\sigma$, and the scalar bias (the parameter to be learned; not to be confused with the estimation bias of an estimator) by $b$, the distribution of $H$ is given by $\pr(H=h) = \pi(h; b) = \sigma((2h-1) b)$ where $h \in \{0,1\}$. After sampling $H$, a scalar reward $R$ is sampled from a distribution conditioned on $H$. We are interested in learning $b$ such that the expected reward $\ex[R]$ is maximized. We also denote the conditional expectation $\ex_{R}[R|H=h]$ by $r(h)$, so $\ex[R]=\ex_{H}[\ex_{R}[R|H]]=\ex[r(H)]$.

Our goal is to compute or estimate $\nabla_b \ex_H[r(H)]$, the gradient of the expected reward w.r.t. parameter $b$, so as to perform gradient ascent on the expected reward. 

A direct computation of the gradient gives:
\begin{align}
	\nabla_b \ex [r(H)] &= 	\nabla_b (\sigma(-b) r(0) + \sigma(b) r(1)) \\
	&= (r(1) - r(0)) \sigma(b) (1 - \sigma(b)). \label{eq:1}
\end{align}
That is, the gradient is positive iff $r(1) > r(0)$ since $\sigma(b) (1 - \sigma(b))$ is always positive. In other words, if the expected reward for firing (i.e. outputting $1$) is larger than the reward for not firing (i.e. outputting $0$), then the unit should fire more by increasing $b$. In fact, we see that the optimal $b^* = \argmax_b \ex[r(H)]$ is infinitely positive if $r(1) > r(0)$  and infinitely negative if $r(1) < r(0)$. However, $r(1) - r(0)$ is generally not known.
%Thus, the sign of the gradient is more important than the exact value of the gradient as we can compute $b^*$ from the sign alone.

%However, unless the unit is an output unit and the reward function (i.e.\ the expected reward conditioned on the activation value of output units) is known, we cannot compute $r(1) - r(0)$ directly. This is because even if we know the reward function, we still need to marginalize over all outgoing units if the unit is a hidden unit.

\subsection{A multi-layer network of Bernoulli-logistic units}

Before discussing the learning rules, we consider some properties of $r(h)$ if the unit is one of the many units in a multi-layer network of Bernoulli-logistic units. In a multi-layer network, units are grouped into different layers. We call the last layer output layer and other layers hidden layers. Units on the output layer are called output units, and units on hidden layers are called hidden units. Activation values of the hidden units are passed to the next layer as inputs, and activation values of the output units are passed to the external environment and determine the reward distribution.

Therefore, if the unit is a hidden unit in a multi-layer network of Bernoulli-logistic units, $H$ is passed to units on next layer to determine their activation values $D_1, D_2, ..., D_m$ by $\pr(D_i=d|H=h) = \sigma((2d - 1) (v_ih + c_i))$ where $v_i$ is the weight connecting from the unit to the next layer's unit $i$, $c_i$ is the bias of the next layer's unit $i$, $d \in \{0, 1\}$, $i \in \{0, 1, ..., m\}$, and $m$ is the number of next layer's units. We call the vector $v=[v_1, ..., v_m]$ \emph{outgoing weight}, and the next layer's units \emph{outgoing units}.

From this, we see that $r(h)$ can be written as some differentiable function $g(v_1h+c_1, ..., v_mh + c_m)$:
\begin{align}
 & r(h) \\
 =& \ex[R|H=h] \\
 =& \sum_{d_{1:m}} \pr(D_{1:m}=d_{1:m}|H=h) \ex[R|D_{1:m}=d_{1:m}, H=h] \\
 =& \sum_{d_{1:m}} \pr(D_{1:m}=d_{1:m}|H=h) \ex[R|D_{1:m}=d_{1:m}] \label{eq:ne0} \\ 
 =& \sum_{d_{1:m}} \sigma((2d_1 - 1) (v_1h + c_1)) ... \sigma((2d_m - 1) (v_mh + c_m)) \ex[R|D_{1:m}=d_{1:m}] \label{eq:ne2} \\
 =:& g(v_1h + c_1, ..., v_mh + c_m), \label{eq:ne1}
\end{align}
where (\ref{eq:ne0}) uses the assumption that $R$ is independent with $H$ when conditioned on $D_{1:m}$ since the network has a multi-layer structure.

Note that for $h \in [0, 1]$, $g$ is a differentiable function of $h$, so $g$ gives a differentiable extension of $r(h)$ to $h \in [0, 1]$. We call this extension a \emph{natural extension} of $r(h)$, which extends $r(h)$ by allowing the unit to pass any values in $[0, 1]$ instead of only $\{0, 1\}$ to its outgoing units. Though differentiable, the natural extension can be highly non-linear, as shown in Figure \ref{fig:1}(a). The natural extension of $r(h)$, which is only applicable for hidden units, is useful as it opens up the possibility of other learning rules that are not possible for output units. For example, we may have some estimates of $r'(0)$ or $r'(1)$ when training the outgoing units, which can be used to approximate $r(1) - r(0)$ by first-order Taylor approximation.

%By Taylor's formula, we can write $r(1) - r(0)$ as (assuming the Taylor series converges to ):
%\begin{align}
%  & r(1) - r(0)\\
%=& g(v_1 + c_1, ..., v_m + c_m) - g(c_1, ..., c_m)\\
%=& Dg(c)(v) + \frac{1}{2!}D^2g(c)(v^2) + \frac{1}{3!}D^3g(c)(v^3) + ... , \label{eq:tf0}
%\end{align}
%where $D^p g(c)(v^p)$ denotes the $p$-order derivative of $g$ evaluated at $c = (c_1, c_2, ..., c_m)$ that takes $(v, v, ..., v)$ (repeated $p$ times) as input. For example, $Dg(c)(v) = v \cdot \nabla g(c)$, the dot product between $v$ and the gradient of $g$ evaluated at $c$, and $D^2g(c)(v^2) = v^T \nabla^2 g(c) v$, where $\nabla^2 g(c)$ denotes the Hessian of $g$ evaluated at $c$. We can also evaluate the Taylor's approximation at $v+c$ instead of $c$:
%\begin{align}
%	& r(1) - r(0)\\
%	=& Dg(v+c)(v) - \frac{1}{2!}D^2g(v+c)(v^2) + \frac{1}{3!}D^3g(v+c)(v^3) + ... . \label{eq:tf1}
%\end{align}
%We note that the $i$ entry of $\nabla g(hv+c)$ is the gradient of $\ex[R|H=h]$ w.r.t. $c_i$.

We are now ready to discuss the five different estimates of $\nabla_b \ex[r(H)]$.

\subsection{REINFORCE}

The gradient can be computed as:
\begin{align}
   \nabla_b \ex [r(H)] &= \ex [r(H)  \nabla_b \log \pr (H)] \\
					   &= \ex [r(H)  (H - \sigma(b))].
\end{align}
Suppose that we have an unbiased estimate of $r(h)$ for any $h$, denoted by $\hat{r}(h)$, we can
then estimate the gradient by:
\begin{align}
	\Delta_{rei}b = \hat{r}(H) (H - \sigma(b)), \label{eq:rei1}
\end{align}
which is an unbiased estimate of the gradient:
 \begin{align}
 	\ex[\Delta_{rei}b] &= \mathbb{E}_{H} [\ex[\hat{r}(H)|H] (H - \sigma(b))] \\
 	&= \ex [r(H)  (H - \sigma(b))] \\
 	&= \nabla_b \ex [r(H)].
 \end{align}

Note that the $R$ sampled (conditioned on $H$) is an unbiased estimate of $r(H)$, so $R (H - \sigma(b))$ is an unbiased estimate of the gradient and this becomes the REINFORCE learning rule \cite{williams1992simple}.

However, if $r(0)$ and $r(1)$ are very close, then the expected value of  $\Delta_{rei}b$ is close to zero as shown by (\ref{eq:1}). Combined with the variance of $\hat{r}(H)$, this makes the sign of $\Delta_{rei}b$ almost random, and so it takes many steps of gradient ascent to reach $b^*$. This is indeed the case when the unit is a hidden unit in a large network of Bernoulli-logistic units.

\subsection{STE Backprop}

Suppose that (i) we can extend $r(h)$ to a differentiable function, and (ii) we have an  estimate of $r'(h)$ for any $h$ ($f'$ denotes the derivative of $f$), denoted by $\hat{r}'(h)$; then we can apply STE backprop \cite{bengio2013estimating}, a common method for training Bernoulli-logistic units ($1\{\cdot \}$ denotes the indicator function and $U$ denotes a uniformly distributed and independent variable):

\begin{align}
	\nabla_b \ex [r(H)] &= \nabla_b \mathbb{E}_U [r(1\{\sigma(b) > U\})] \label{eq:sbp1} \\
	& = \mathbb{E}_U [\nabla_b  r(1\{\sigma(b) > U\})]  \\
	& = \mathbb{E}_U [r'(1\{\sigma(b) > U\})  \nabla_b 1\{\sigma(b) > U\} ]  \\	
	& \approx  \mathbb{E}_U  [r'(1\{\sigma(b) > U\})  \nabla_b \sigma(b) ]  \label{eq:sbp2} \\
	& = \mathbb{E}_U  [r'(1\{\sigma(b) > U\})  \sigma(b)(1-\sigma(b)) ] \\
	& = \ex[r'(H)]\sigma(b)(1-\sigma(b)) \label{eq:sbp3},
\end{align}
where (\ref{eq:sbp1}) uses the fact that $1\{\sigma(b) > U\}$ has the same distribution as $H$, and (\ref{eq:sbp2}) uses the approximation of $\nabla_x 1\{x > u\} \approx 1$ in STE backprop (note that the actual derivative of $f(x) = 1\{x > u\}$ is zero at $x \neq u$ and does not exist at $x=u$).

\begin{figure}%
	\centering
	\subfloat[\centering $r(h)$ from natural extension]{{\includegraphics[width=0.5\textwidth]{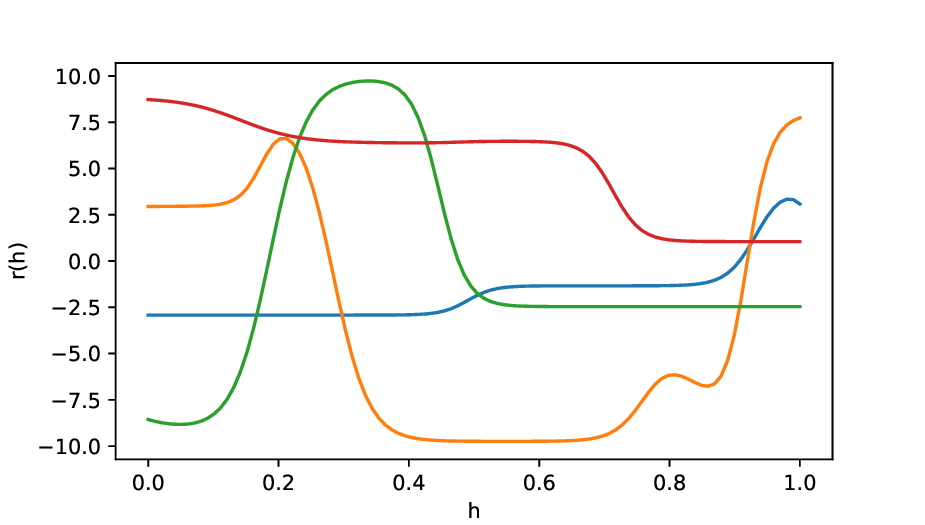} }}%
	\subfloat[\centering Two different extended $r(h)$]{{\includegraphics[width=0.5\textwidth]{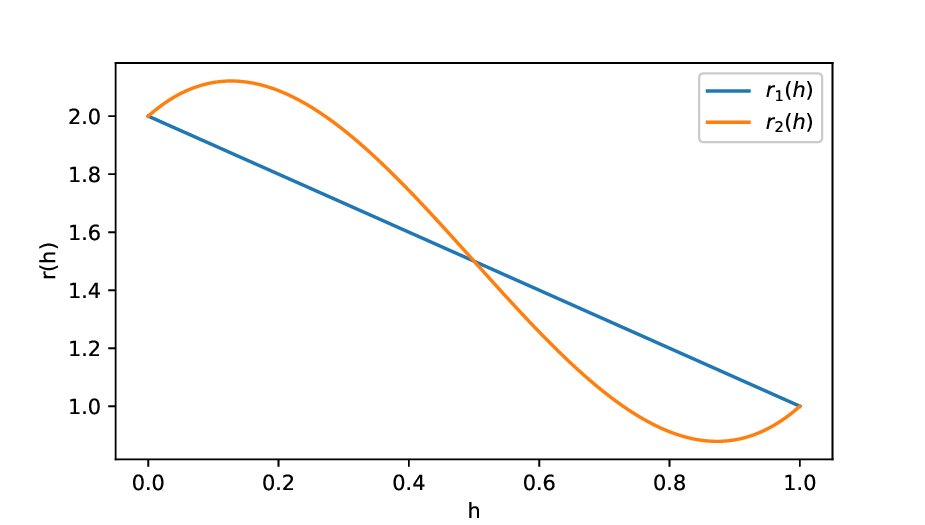} }}%
	\caption{(a) The natural extension of $r(h)$ in the form of (\ref{eq:ne2}). 8 outgoing units ($m=8$) is used and all parameters ($v, c, \ex[R|D_{1:m}=d_{1:m}]$) are randomly sampled. Four resulting $r(h)$ are shown on the graph. (b) If the extension is linear (as shown in $r_1(h)$), then the approximation of $r(1) - r(0)$ by $r'(0)$ or $r'(1)$ is exact. But if the extension is non-linear (as shown in $r_2(h)$), then the approximation of $r(1) - r(0)$ by $r'(0)$ or $r'(1)$ can have a wrong sign.}%
	\label{fig:1}%
\end{figure}

Thus, the estimate of the gradient by STE backprop is given by:
\begin{align}
	\Delta_{ste}b = \hat{r}'(H) \sigma(b)(1-\sigma(b)). \label{eq:sbp4}
\end{align}
We observe that the form of $\Delta_{ste}b$ is very similar to (\ref{eq:1}). Intuitively, STE backprop is using $\hat{r}'(H)$, i.e.\ the estimated derivative of $r$ at $h=H$, to approximate $r(1) - r(0)$. If $r$ is a linear function and $\hat{r}'(H)$ is unbiased, then $\ex[\hat{r}'(H)] = \ex[r'(H)] = r(1) - r(0)$ and the estimate $\Delta_{ste}b$ is unbiased. However, if $r$ is a non-linear function, it is possible that $\Delta_{ste}b$ always has the wrong sign and thus $b$ converges to a value that \emph{minimizes} $\ex[R]$ instead. This is illustrated in Figure \ref{fig:1}(b): in the second $r(h)$, the slope at $h=0$ or $h=1$ are both positive, so the approximated $r(1) - r(0)$ is positive. However, the true $r(1) - r(0)$ is negative, and $\Delta_{ste}b$ always has a wrong sign in this case.

The reason for this approximation error is that increasing $b$ by a small amount increases the weight placed (or the probability) at $r(1)$ and decreases the weight placed (or the probability) at $r(0)$ by a small amount, instead of increasing $h$ by a small amount (we cannot `move' on the curve $r(h)$). The latter case corresponds to $H = \sigma(b)$, i.e. the activation value of the unit is deterministic and equals $\sigma(b)$. In fact, backprop applied on a deterministic ANN with sigmoid activation function gives almost the same parameter update as STE backprop applied on an ANN of Bernoulli-logistic units. In other words, STE backprop treats the network as if the hidden units are outputting expected values instead of sampled values.

\subsubsection{Natural Extension}

The two assumptions of having an extension and an estimate $\hat{r}'(h)$ can be satisfied by using the natural extension $g$ defined in (\ref{eq:ne1}) if the unit is a hidden unit in a network. To estimate $r'(h)$, we observe:
\begin{align}
r'(h) &= \nabla_h g(v_1h + c_1, v_2h+ c_2, ..., v_mh + c_m) \\
&= \sum_{i=1}^m v_i \nabla_{i} g(v_1h + c_1, v_2h+ c_2, ..., v_mh + c_m) \\
&= \sum_{i=1}^m v_i  \nabla_{c_i} \ex [R|H=h]. 
\end{align}

Assuming the outgoing units are also learning through adding some estimates of $\nabla_{c_i} \ex [R|H]$ to its bias $c_i$, we can use the change of $c_i$ as an estimate of $\nabla_{c_i} \ex [R|H]$ and plug it in the above formula to obtain $\hat{r}'(H)$. The estimate of $\nabla_{c_i} \ex [R|H]$ for outgoing units can be obtained by STE backprop if the outgoing units are hidden units, and REINFORCE (\ref{eq:rei1}) or direct computation (\ref{eq:1}) if the outgoing units are output units.

However, the natural extension of $r$ is not linear in most cases, so STE backprop gives a biased estimate. From the Taylor expansion of $g$ in (\ref{eq:ne1}), the estimation bias is in the order of $\vert\vert v \vert\vert^2$, the squared norm of the outgoing weight. Note that this estimation bias accumulates across layers as $\hat{r}'(H)$ is no longer an unbiased estimator of $r'(H)$ for units with hidden outgoing units.

\subsection{Weight Maximization} \label{sec:wm}

Both the parameter updates given by REINFORCE and STE backprop are not necessarily zero when the unit is not firing. A desirable property of a learning rule is that the unit's parameters are not updated when the unit is not firing. This can be achieved by using $r(0)$ as a baseline in REINFORCE:
\begin{align}
	\nabla_b \ex [r(H)] &= \ex [(r(H) -r(0)) (H - \sigma(b))].
\end{align}
Note that $(R - r(0))H$ is an unbiased estimate of $r(H) -r(0)$ conditioned on $H$: if $H=0$, then $\ex[(R - r(0))H|H=0] = 0 = r(H) - r(0)$; if $H=1$, then $\ex[(R - r(0))H|H=1] = \ex[R|H=1] - r(0) = r(H) - r(0)$. Thus, an unbiased estimate of the gradient is:
\begin{align}
	\Delta b = (R - r(0))H (H - \sigma(b)) \label{eq:rei2}.
\end{align}
However, the above estimate cannot be computed since $r(0)$ is not known in general. Suppose that (i) we can extend $r(h)$ to a differentiable function, and (ii) we have an  estimate of $r'(h)$ for any $h$, denoted as $\hat{r}'(h)$, then we may approximate $(R - r(0))H$ by $\hat{r}'(H)H$, leading to the following estimate of the gradient:
\begin{align}
	\Delta_{wm} b = \hat{r}'(H)H (H - \sigma(b)) \label{eq:wm1}.
\end{align}
Compared to REINFORCE, Weight Maximization replaced the global reward $R$ by the individual reward $\hat{R} := \hat{r}'(H)H$, which equals zero when the unit is not firing and equals an approximation of $r(1) - r(0)$ by $r'(1)$ when the unit is firing. Similar to STE backprop, this approximation requires $r$ to be linear, which does not hold in most cases.

\subsubsection{Natural Extension}

The two assumptions of having an extension and an estimate $\hat{r}'(h)$ can be satisfied by using the natural extension defined in (\ref{eq:ne1}) when the unit is a hidden unit in a network. We can estimate $r'(h)h$ by:
\begin{align}
	r'(h)h &= h \nabla_h g(v_1h + c_1, v_2h+ c_2, ..., v_mh + c_m) \\
	&= \sum_{i=1}^m v_i h \nabla_{i} g(v_1h + c_1, v_2h+ c_2, ..., v_mh + c_m) \\
	&= \sum_{i=1}^m v_i  \nabla_{v_i} \ex [R|H=h]. 
\end{align}
Assuming the outgoing units are also learning through adding some estimates of $\nabla_{v_i} \ex [R|H]$ to its weight $v_i$, we can use the change of $v_i$ as an estimate of $\nabla_{v_i} \ex [R|H]$ and plug it in the above formula to obtain $\hat{r}'(H)H$. This leads to the alternative perspective of maximizing outgoing weights in Weight Maximization \cite{chung2022learning}.

However, the natural extension of $r$ is not linear in most cases, so Weight Maximization gives a biased estimate. From the Taylor expansion of $g$ in (\ref{eq:ne1}), the estimation bias is in the order of $\vert\vert v \vert\vert^2$, the squared norm of the outgoing weight. Note that this estimation bias accumulates across layers as $\hat{r}'(H)$ is no longer an unbiased estimator of $r'(H)$ for units with hidden outgoing units.

\subsection{High-order Weight Maximization} \label{sec:hwm}

Both STE backprop and Weight Maximization assume the extended $r$ to be linear. In particular, Weight Maximization uses ${r}'(1)$ to estimate $r(1) - r(0)$, which is a first-order Taylor approximation. We may use $p$-order Taylor approximation to get a better estimate (where $p \geq 1$):
\begin{align}
	r(1) - r(0) \approx r^{(1)}(1) - \frac{1}{2!} r^{(2)}(1) +  \frac{1}{3!} r^{(3)}(1) ... + (-1)^{p+1} \frac{1}{p!} r^{(p)}(1), 
\end{align}
where $r^{(p)}(h)$ denotes the $p$-order derivative of $r$. With this new approximation, (\ref{eq:wm1}) is generalized to:
\begin{align}
	\Delta_{wm-p} b = \left(\sum_{k=1}^p  (-1)^{k+1} \frac{1}{k!}  \hat{r}^{(k)}(H) \right) H (H - \sigma(b)) \label{eq:wm2}.
\end{align}
This gives the parameter update for $p$-order Weight Maximization, which replaces the global reward $R$ in REINFORCE by the individual reward:
\begin{align}
	\hat{R} := \left(\sum_{k=1}^p  (-1)^{k+1} \frac{1}{k!}  \hat{r}^{(k)}(H) \right) H.
\end{align}
We call the learning rule of updating $b$ by $\Delta_{wm-p} b$ in $(\ref{eq:wm2})$ \emph{$p$-order Weight Maximization}. Note that first-order Weight Maximization is equivalent to Weight Maximization discussed in Section \ref{sec:wm}. 

Computing $\hat{R}$ requires estimates of  ${r}^{(k)}(h)$ for $k=1, 2, ..., p$. We discuss how to obtain these estimates next.

\subsubsection{Natural Extension}

Again, we use the natural extension of $r$. Let first consider ${r}^{(1)}(h)$:
\begin{align}
	&r^{(1)}(h) \\
	=& \nabla_h \ex[R|H=h]\\
	=& \nabla_h \sum_{d_{1:m}} \pr(D_{1:m}=d_{1:m}|H=h) \ex[R|D_{1:m}=d_{1:m}] \\
	=& \sum_{d_{1:m}} \pr(D_{1:m}=d_{1:m}|H=h) \ex[R|D_{1:m}=d_{1:m}]  \nabla_h \log \pr(D_{1:m}=d_{1:m}|H=h) \\
	=& \ex [R \nabla_h \log \pr(D_{1:m}|H=h)|H=h]   \\	
	=& \ex \left[\sum_{i=1}^m R  v_i (D_i - \sigma(v_ih+c_i))\bigg|H=h\right], \label{eq:ho1}
\end{align}
Thus, an unbiased estimate of ${r}^{(1)}(h)$ is $\sum_{i=1}^m R v_i (D_i - \sigma(v_ih+c_i))$. If the outgoing units are hidden units and are trained by Weight Maximization, each unit receives a different reward $\hat{R}_i$ that can be used to replace the global reward $R$ in the estimate. If the outgoing units are output unit, we let $\hat{R}_i=R$. Therefore, we have\footnote{For simplicity let first assume $\nabla_{c_i} \ex [\hat{R}_i|H=h] = \nabla_{c_i} \ex [R|H=h]$; i.e.\ we ignore the approximation error of Weight Maximization applied on outgoing units. We can still use the global reward $R$ if the outgoing units are hidden units, but this leads to a higher variance.}:
\begin{align}	
\ex \left[\sum_{i=1}^m R  v_i (D_i - \sigma(v_ih+c_i))\bigg|H=h\right] = \ex \left[\sum_{i=1}^m \hat{R}_i  v_i (D_i - \sigma(v_ih+c_i))\bigg|H=h\right].
\end{align}
Note that $\hat{R}_i (D_i - \sigma(v_ih+c_i))$ is also the update to weight $v_i$ for outgoing unit $i$. To simplify notation, we define the following terms:
\begin{align}
	s^{(k)} = \begin{cases}
		\sum_{i=1}^m \hat{R}_i v_i (D_i -\sigma(v_ih+c_i)), & \text{if $k = 1$}\\
		\sum_{i=1}^m \hat{R}_i v_i^k (-\sigma^{(k)}(v_ih+c_i)), & \text{if $k = 2, 3, ...$} \label{eq:hwm1}
	\end{cases}
\end{align}	
\begin{align}
	t^{(k)} = \begin{cases}
		\sum_{i=1}^m v_i (D_i -\sigma(v_ih+c_i)), & \text{if $k = 1$}\\
		\sum_{i=1}^m v_i^k (-\sigma^{(k)}(v_ih+c_i)). & \text{if $k = 2, 3, ...$}
	\end{cases} \label{eq:hwm2}
\end{align}	
We see that $\hat{r}^{(1)}(h) \defeq s^{(1)}$ is an unbiased estimator of ${r}^{(1)}(h)$. Note that $\nabla_h s^{(k)} = s^{(k+1)}$ and $\nabla_h t^{(k)} = t^{(k+1)}$. Next, consider ${r}^{(2)}(h)$:
\begin{align}
	r^{(2)}(h) &= \nabla_h r^{(1)}(h) \\
	&= \nabla_h \ex[s^{(1)}|H=h] \\
	&= \ex[ \nabla_h s^{(1)} + s^{(1)} \nabla_h \log \pr(D_{1:m}|H=h)|H=h] \\
	&= \ex[s^{(2)} + s^{(1)} t^{(1)}|H=h].
\end{align}
Therefore, $\hat{r}^{(2)}(h) \defeq s^{(2)} + s^{(1)} t^{(1)}$ is an unbiased estimator of $r^{(2)}(h)$. Next, consider ${r}^{(3)}(h)$:
\begin{align}
	r^{(3)}(h) &= \nabla_h r^{(2)}(h) \\
	&= \nabla_h \ex[s^{(2)} + s^{(1)} t^{(1)}|H=h] \\
	&= \ex[ \nabla_h (s^{(2)} + s^{(1)} t^{(1)}) + (s^{(2)} + s^{(1)} t^{(1)})t^{(1)}|H=h] \\
	&= \ex[s^{(3)} + 2 s^{(2)} t^{(1)} + s^{(1)} t^{(2)} + s^{(1)} (t^{(1)})^2 |H=h].
\end{align}
Therefore, $\hat{r}^{(3)}(h) \defeq s^{(3)} + 2 s^{(2)} t^{(1)} + s^{(1)} t^{(2)} + s^{(1)} (t^{(1)})^2$ is an unbiased estimator of $r^{(3)}(h)$. We can continue to compute an unbiased estimator of $r^{(4)}(h), r^{(5)}(h)$ and so on in the same way. See Appendix \ref{sec:a1} for the general formula of an unbiased estimator of $r^{(p)}(h)$. These estimates are then plugged in (\ref{eq:wm2}) to compute $\hat{R}$.

If the high-order derivatives of $g$ in (\ref{eq:ne2}) are bounded, i.e.\ there exists $B$ such that $|\partial_{i_1} \partial_{i_2} ... \partial_{i_n} g(hv+c)| < B$ for any $h \in [0, 1]$, $i_j \in \{1, 2, ..., m\}$ for $j \in \{1, 2, ..., n\}$, $n > 0$, then
\begin{align}
 r(1) - r(0) &= \ex \left[\sum_{k=1}^p  (-1)^{k+1} \frac{1}{k!}  \hat{r}^{(k)}(1) \bigg|H=1 \right] + \mathcal{O}\left(\frac{\vert\vert v \vert \vert ^{p+1} }{(p+1)!} \right).
\end{align}
That is, the estimation bias is in the order of $\frac{\vert\vert v \vert \vert ^{p+1} }{(p+1)!}$, which converges to $0$ as $p \rightarrow \infty$. The proof is based on the error bound of Taylor approximation applied on $g$.

\subsubsection{Unbounded Derivatives}

\begin{figure}%
	\centering
	\subfloat[\centering $\sigma^{(p)}(x)$ for different $p$]{{\includegraphics[width=0.5\textwidth]{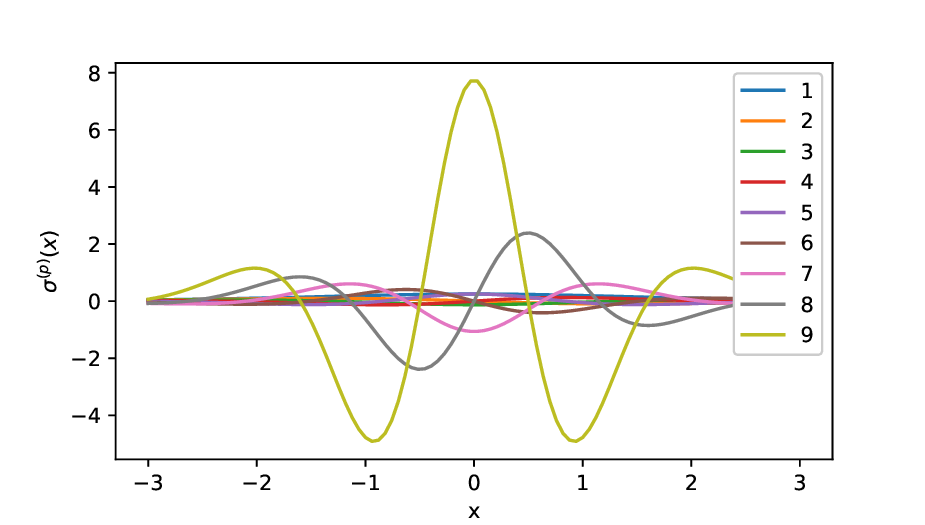} }}%
	\subfloat[\centering Absolute error of $p$-order Taylor approximation]{{\includegraphics[width=0.5\textwidth]{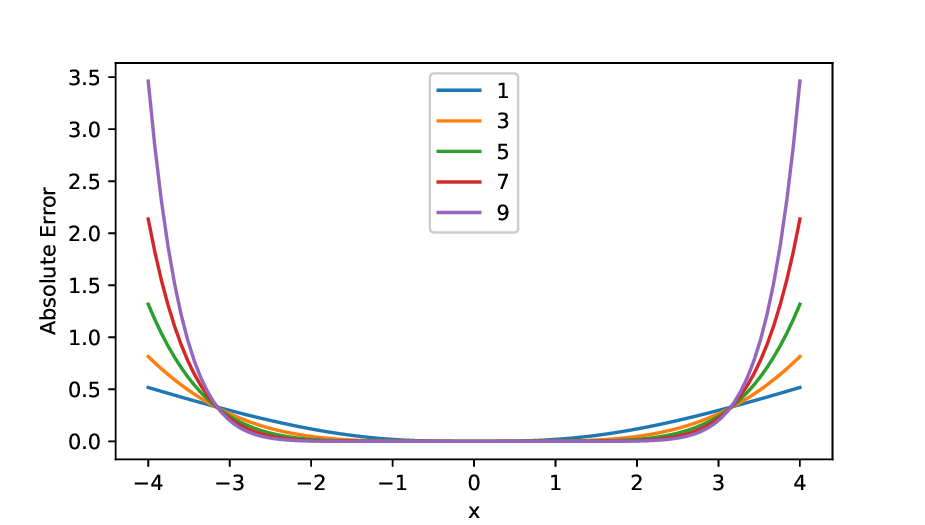} }}%
	\caption{(a) The high-order derivatives of $\sigma(x)$ are unbounded. Observe that the magnitude of $\sigma^{(p)}(x)$ grows with $p$. (b) Absolute error of $p$-order Taylor approximation of $\sigma(x)$ at $x=0$. Observe that a higher-order Taylor approximation leads to a \emph{larger} error when $|x| \geq \pi$.}%
	\label{fig:2}%
\end{figure}

Unfortunately, \emph{the high-order derivatives of $g$ are not bounded for a network of Bernoulli-logistic units}. This is because of the sigmoid function in $g$ as shown in (\ref{eq:ne2}), and the $p$-order derivatives of the sigmoid function are unbounded (see Fig \ref{fig:2}). For example, estimating $\sigma(4)$ at $x=0$ by Taylor approximation results in a diverging sequence as $p \rightarrow \infty$. Denoting $C_{p+1} := \max_{x} |\sigma^{(p+1)}(x)|$, the estimation bias of $p$-order Weight Maximization applied on a network of Bernoulli-logistic units is  $\mathcal{O}\left(\frac{C_{p+1}\vert\vert v \vert \vert ^{p+1} }{(p+1)!}\right)$, which is unbounded as $p \rightarrow \infty$ if $\vert\vert v \vert\vert \geq \pi$ (see Appendix \ref{sec:a2}).

 %This makes the estimation bias being in the order of $\vert\vert v \vert \vert ^{p+1}$ (the $(p+1)!$ cancels out), which converges to $0$ iff $\vert\vert v \vert \vert < 1$ and diverges iff $\vert\vert v \vert \vert > 1$. 

In experiments, we observe that high-order Weight Maximization performs similar to low-order Weight Maximization in the beginning of training. But as the norm of outgoing weight grows large during training, the estimation bias becomes \emph{larger} for a higher $p$, and hence high-order Weight Maximization's performance deteriorates quickly in the middle of training.

Maybe using activation functions other than the sigmoid function or using weight decay to prevent the outgoing weight's norm from growing too large can prevent this issue. But there is another method to tackle this issue, which will be discussed next.

%\textcolor{blue}{Note: the failure of high-order Weight Maximization stems from the divergent Taylor series for the sigmoid function. Maybe Fourier series can be explored instead of Taylor series.}

\subsection{Unbiased Weight Maximization} \label{sec:uwm}
Let us consider Fig \ref{fig:1} again. All previous methods (except REINFORCE) are based on using the derivative of $r$ to approximate $r(1) - r(0)$. STE backprop approximates it by $r'(0)$ or $r'(1)$; Weight Maximization approximates it by $r'(1)$; high-order Weight Maximization approximates it by high-order Taylor series of $r$ at $h=1$. The former two methods assume the $r(h)$ to be linear, which is not true in general; the last method may lead to a divergent Taylor series if the outgoing weight is too large. Are there any other simple methods to approximates $r(1) - r(0)$ based on $r'(h)$?

Fundamental theorem of calculus tells us that we can integrate $r'(h)$ over $h$ to get $r(1) - r(0)$:
\begin{align}
	\int^1_0 r'(u) du = r(1) - r(0).
\end{align}
This can in turn be written as:
\begin{align}
	\ex [r'(U)] = r(1) - r(0),
\end{align}
where $U$ is an independent and uniformly distributed random variable. Thus, if we have an unbiased estimate of $r'(u)$ for any $u \in [0,1]$, denoted as $\hat{r}'(u)$, we may use the following estimate instead:
\begin{align}
	\Delta_{uwm} b = \hat{r}'(U)H (H - \sigma(b)) \label{eq:rp1}
\end{align}
which is an unbiased estimator of $\nabla_b \ex[r(H)]$:
\begin{proof}	
\begin{align}	
	\ex	[\Delta_{uwm} b] &= \ex [\hat{r}'(U)H (H - \sigma(b))] \\
	&= \ex [\hat{r}'(U)H (H - \sigma(b))|H=1] \pr(H=1) \\
	&= \ex [\hat{r}'(U) (1 - \sigma(b))|H=1] \pr(H=1) \\	
	&= \ex [\hat{r}'(U) |H=1] (1 - \sigma(b)) \sigma(b) \\		
	&= \ex [{r}'(U)] (1 - \sigma(b)) \sigma(b) \\		
    &= (r(1) - r(0)) (1 - \sigma(b)) \sigma(b) \\				
    &= \nabla_b \ex[r(H)].    
\end{align}
The last line follows from (\ref{eq:1}).
\end{proof}
Note that $\Delta_{uwm} b$ is very similar to $\Delta_{wm} b$ in (\ref{eq:wm1}) - we only replaced $\hat{r}'(H)$ in (\ref{eq:wm1}) with $\hat{r}'(U)$, i.e.\ we evaluate the gradient at a random point on $[0,1]$ instead of $H$. 

We also define the individual reward $\hat{R} := \hat{r}'(U)H$, which is an unbiased estimator of $r(H) - r(0)$ conditioned on $H$:
\begin{align}
	\ex[\hat{R}|H] &= \ex[\hat{r}'(U)H|H] \\
				   &= \begin{cases}
				       0, \text{if $H = 0$}\\
					   \ex[\hat{r}'(U)], \text{if $H = 1$}
   					  \end{cases}	\\
   				  &= r(H) - r(0). \label{eq:rp1.5}
\end{align}

We call the learning rule of updating $b$ by $\Delta_{uwm} b$ in $(\ref{eq:rp1})$ \emph{unbiased Weight Maximization}.

\subsubsection{Natural Extension}
The only remaining question is how can we obtain an unbiased estimate of $r'(u)$ for any $u \in [0,1]$ when the unit can only output $\{0, 1\}$. Again, let consider the natural extension of $r$. By importance sampling, we have, for any $u \in [0,1]$:
\begin{align}
	\ex \left[\frac{\pr(D_{1:m}=d_{1:m}|H=u)}{\pr(D_{1:m}=d_{1:m}|H)} \sum_{i=1}^m \hat{R}_i \nabla_u \log \pr(D_i=d_i|H=u)  \right] = {r}'(u),
\end{align}
where $\hat{R}_i$ denotes the individual reward to the outgoing unit $i$, which equals the global reward $R$ if the outgoing unit is an output unit, or the individual reward computed by unbiased Weight Maximization applied on the outgoing unit if the outgoing unit is a hidden unit. We denote the term in the expectation by $\hat{r}'(u)$, which is an unbiased estimator of ${r}'(u)$ and gives an iterative formula to compute the individual reward to all hidden units in a network.
\begin{proof}
\begin{align}	
 & \ex \left[\frac{\pr(D_{1:m}=d_{1:m}|H=u)}{\pr(D_{1:m}=d_{1:m}|H)} \sum_{i=1}^m \hat{R}_i  \nabla_u \log \pr(D_i=d_i|H=u)  \right] \\
=& \sum_{h, d_{1:m}} \pr(H=h) \pr(D_{1:m}=d_{1:m}|H=u) \sum_{i=1}^m \ex[\hat{R}_i|D_{1:m}=d_{1:m}, H=h]  \nabla_u \log \pr(D_i=d_i|H=u) \label{eq:rp2}
\end{align}
If the outgoing units are output units, then $\hat{R}_i = R$ by definition, and (\ref{eq:rp2}) becomes:
\begin{align}	
& \sum_{d_{1:m}} \pr(D_{1:m}=d_{1:m}|H=u) \sum_{i=1}^m \ex[{R}|D_{1:m}=d_{1:m}] \log  \nabla_u \pr(D_i=d_i|H=u) \label{eq:rp3}\\
=& \sum_{d_{1:m}} \pr(D_{1:m}=d_{1:m}|H=u) \ex[{R}|D_{1:m}=d_{1:m}] \sum_{i=1}^m  \log  \nabla_u \pr(D_i=d_i|H=u) \\
=& \nabla_u  \sum_{d_{1:m}} \pr(D_{1:m}=d_{1:m}|H=u) \ex[{R}|D_{1:m}=d_{1:m}] \\
=& \nabla_u \ex[{R}|H=u] \\
=& \hat{r}'(u), 
\end{align}
where $(\ref{eq:rp3})$ uses the assumption that $R$ is independent with $H$ conditioned on $D_{1:m}=d_{1:m}$ since the network has a multi-layer structure.

If the outgoing units are hidden units, then $\hat{R}_{i}$ is computed by unbiased Weight Maximization applied on the outgoing units, and $\ex[\hat{R}_{i}|D_i=d_i, H=h] = \ex[R|D_i=d_i, H=h] - \ex[R|D_i=0, H=h]$ by (\ref{eq:rp1.5}). So (\ref{eq:rp2}) becomes:
\begin{align}
&\sum_{d_{1:m}} \pr(D_{1:m}=d_{1:m}|H=u) \sum_{i=1}^m \ex[{R}|D_{1:m}=d_{1:m}] \log  \nabla_u \pr(D_i=d_i|H=u) - \nonumber \\
&\sum_{d_{1:m}} \pr(D_{1:m}=d_{1:m}|H=u) \sum_{i=1}^m \ex[{R}|D_i=0; D_{1:m \backslash i}=d_{1:m \backslash i}, H=h] \log  \nabla_u \pr(D_i=d_i|H=u) 	
\end{align}	
It suffices to show that the second term is zero since the first term is (\ref{eq:rp3}):
\begin{align}
	&\sum_{d_{1:m}} \pr(D_{1:m}=d_{1:m}|H=u) \sum_{i=1}^m \ex[{R}|D_i=0; D_{1:m \backslash i}=d_{1:m \backslash i}] \log  \nabla_u \pr(D_i=d_i|H=u) 	\\
	=& \sum_{d_{1:m\backslash i}}\pr(D_{1:m\backslash i }=d_{1:m\backslash i }|H=u) \sum_{i=1}^m \ex[{R}|D_i=0; D_{1:m \backslash i}=d_{1:m \backslash i}, H=h] \nabla_u \sum_{d_i} \pr(D_i=d_i|H=u)  \\
	=& 0.
\end{align}	
	
Note that the event of $H=u$ is not well defined since $H$ can only take binary values, and so $\pr(D_{1:m}=d_{1:m}|H=u)$ is defined by allowing $H$ to pass any values in $[0,1]$ to outgoing units (i.e.\ extending $h$ to $[0,1]$ in (\ref{eq:ne2})).
\end{proof}

$\hat{r}'(u)$ can then be used to compute $\hat{R}$ by $\hat{r}'(u)H$, and $\Delta_{uwm}b$ can be computed to update $b$. 

Note that we can also use multiple i.i.d. $U_1, U_2, ... U_M$ or numerical integration\footnote{Analytical integration is possible iff each outgoing unit receives the same reward.} to estimate $\ex[\hat{r}'(U)]$ to reduce variance, but it does not give a better performance in our experiments.

\subsection{Relationship between different Weight Maximization}

We have discussed three different forms of Weight Maximization: Weight Maximization in Section \ref{sec:wm}, high-order Weight Maximization in Section \ref{sec:hwm}, and unbiased Weight Maximization in \ref{sec:uwm}. The only difference between these algorithms is how the individual reward $\hat{R}$ is computed, as shown in Table \ref{table:1}. This individual reward $\hat{R}$ is used to replace the global reward $R$ in the REINFORCE learning rule.

Weight Maximization is equivalent to $p$-order Weight Maximization when $p=1$, as discussed previously. Weight Maximization is also equivalent to unbiased Weight Maximization if we set $U=H$ instead of uniformly sampling $U$ from $[0,1]$. In other words, Weight Maximization evaluates the gradient $r'(h)$ at endpoints while unbiased Weight Maximization evaluates the gradient $r'(h)$ at a random point in $[0,1]$.

In addition, it can be shown that if the Taylor series of $r$ converges, as $p \rightarrow \infty$, the update given by $p$-order Weight Maximization converges to unbiased Weight Maximization that directly computes $\ex[\hat{r}'(U)]$ instead of estimating it with $\hat{r}'(U)$. The proof is quite long and is omitted here. From this we see that Weight Maximization can also be seen as the first-order Taylor approximation of unbiased Weight Maximization.

Though we include `Weight Maximization' in these learning rules to emphasize their origins, it should be noted that these extensions can no longer be seen as maximizing the norm of outgoing weight as the individual rewards are computed differently compared to Weight Maximization.

\begin{table}
	\caption{Comparison of how individual reward $\hat{R}$ is computed and estimation bias.}
	\label{table:1}
	\centering
	\begin{tabular}{lcc}
		\toprule
		& Formula for $\hat{R}$ & Estimation Bias\\
		\midrule
		REINFORCE & $R$ & 0 \\		
		Weight Max & $\sum_{i=1}^m \hat{R}_i H \nabla_H \log \pr(D_i=d_i|H)$ & $\mathcal{O}\left(\vert\vert v \vert \vert^2\right)$ \\
		$p$-order Weight Max & $\left(\sum_{k=1}^p  (-1)^{k+1} \frac{1}{k!}  \hat{r}^{(k)}(H) \right) H$ where $\hat{r}^{(k)}$ is defined by (\ref{eq:a11}) & $\mathcal{O}\left(\frac{C_{p+1}\vert\vert v \vert \vert ^{p+1} }{(p+1)!}\right)$ \\		
	Unbiased Weight Max & $\frac{\pr(D_{1:m}=d_{1:m}|H=U)}{\pr(D_{1:m}=d_{1:m}|H)} \sum_{i=1}^m \hat{R}_i H \nabla_U \log \pr(D_i=d_i|H=U)$ & 0 \\
		\bottomrule
	\end{tabular}
\end{table}

\section{Experiments} \label{sec:ex}

To test the above algorithms, we consider the $k$-bit multiplexer task. The input is sampled from all possible values of a binary vector of size $k + 2^k$ with equal probability. The output set is $\{-1, 1\}$, and we give a reward of $+1$ if the network's output equals the multiplexer's output and $-1$ otherwise. We consider $k=4$ here, so the dimension of the input space is 20. We call a single interaction of the network from receiving an input to receiving the reward an episode; the series of episodes as the network's performance increases during training is called a run.

We used a three-layer network of Bernoulli-logistic units, with the first and the second hidden layers having $N \in \{8, 16, 32, 48, 64, 96\}$ units and the output layer having a single unit. The update step size for gradient ascent is chosen as $0.005$, and the batch size is chosen as $16$ (i.e.\ we estimated the gradient for $16$ episodes and averaged them in the update). We used the Adam optimizer \cite{kingma2014adam}. These hyperparameters are chosen prior to the experiments without any tuning.

\begin{figure}[h!!!]
	\centering
	\includegraphics[width=\textwidth]{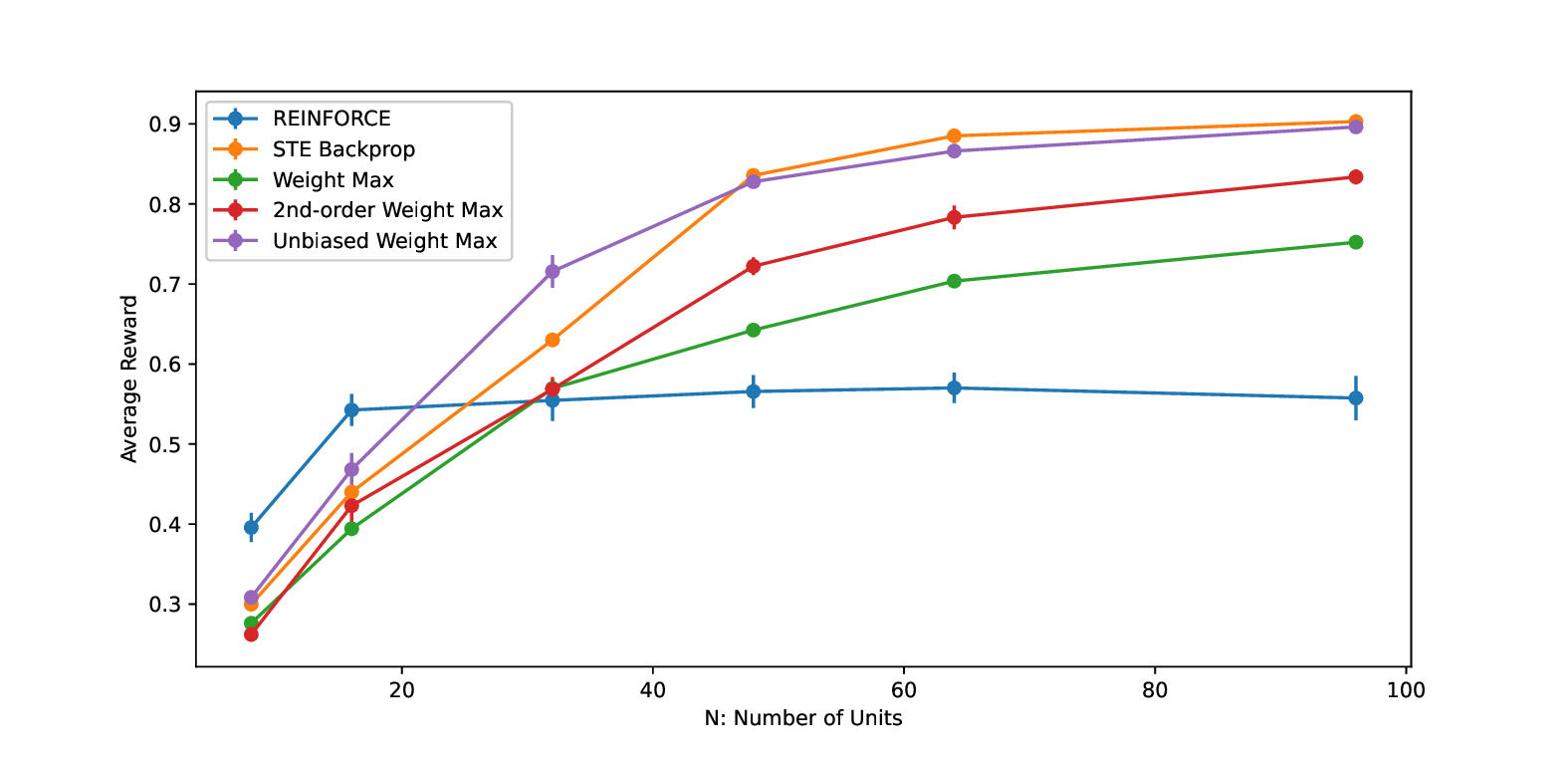}
	\caption{Average rewards of the first $5e6$ episodes in the multiplexer task with a different $N$. Results are averaged over 5 independent runs, and the error bar represents the standard deviation over the runs.}
	\label{fig:3}
\end{figure}

\begin{figure}[h!!!]
	\centering
	\includegraphics[width=\textwidth]{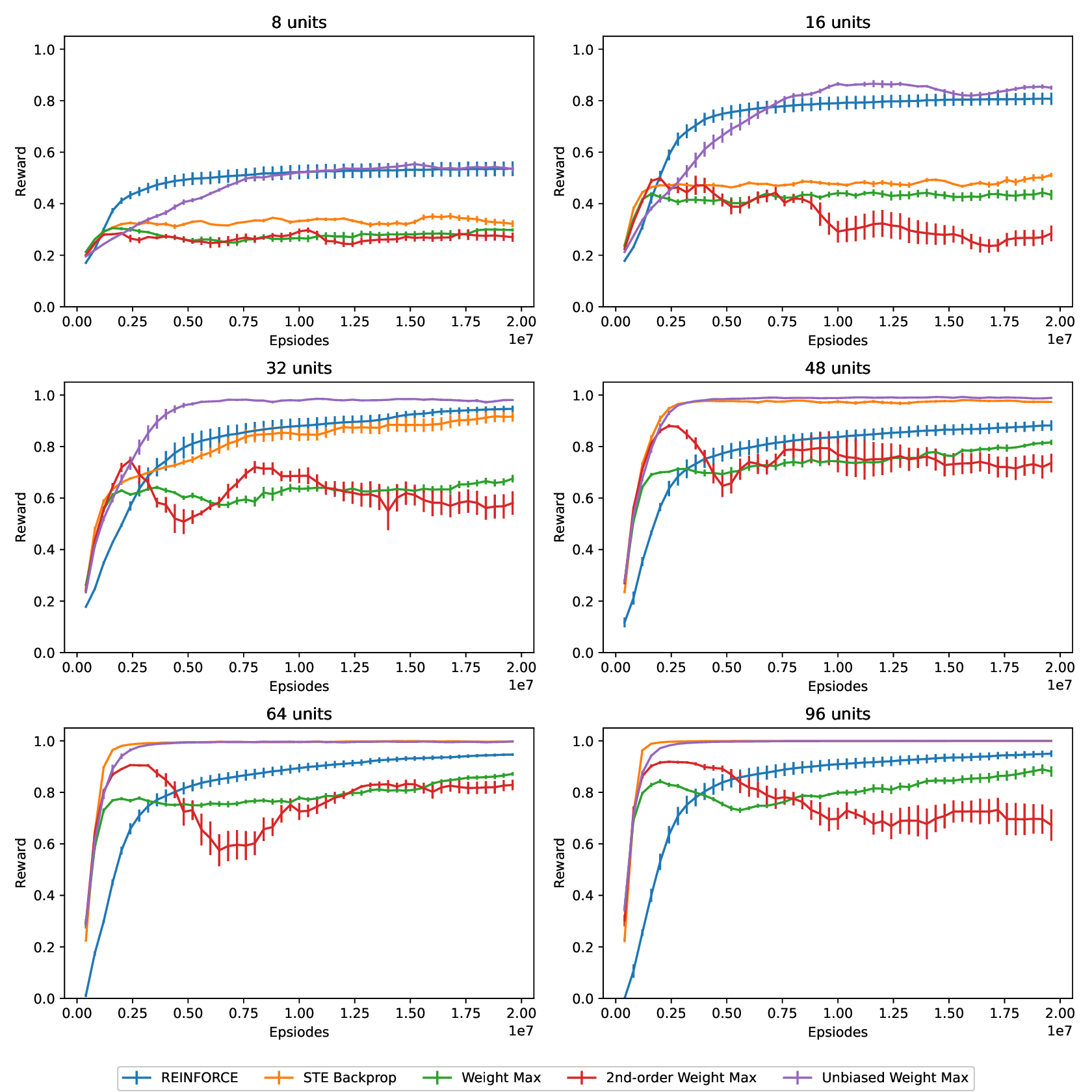}
	\caption{Running average of rewards in each episode throughout a run in the multiplexer task with a different $N$. Results are averaged over 5 independent runs, and the error bar represents the standard deviation over the runs.}
	\label{fig:4}
\end{figure}

The average reward in the first $5e6$ episodes for networks with different $N$ and learning rules is also shown in Figure \ref{fig:3}. The learning curve (i.e. the reward of each episode throughout a run) for each $N$ is also shown in Figure \ref{fig:4}. Note that the results are averaged over 5 independent runs, and the error bar represents the standard deviation over runs.

We observe that:
\begin{itemize}
	\item REINFORCE: When $N=8$, REINFORCE gives the best performance in terms of both learning speed and asymptotic performance - it is easy to correlate units' activation values with the reward when the network size is small; but as $N$ grows larger than 16, the learning curve does not change much - REINFORCE does not benefit from a larger network due to the increased noise.
	\item STE Backprop: When $N$ is small, STE backprop has a poor asymptotic performance - when there are only a few units, the norm of weight has to be large, and so the network has a very different dynamic than an ANN with deterministic and sigmoid units, which is essentially the assumption of STE backprop. But when the network is large, the dynamic becomes similar, and so STE backprop works well when $N$ is large.
	\item Weight Maximization and second-order Weight Maximization: Both learn fast in the first 1e6 episodes, but performance quickly deteriorates afterward. As the norm of weight grows large during training, estimation bias also grows large. Note that weight decay is required to make Weight Maximization work well, but we did not use weight decay in our experiments as it requires extensive tuning. 
	\item Unbiased Weight Maximization: It works well in terms of both learning speed and asymptotic performance under all $N$. The unbiased property makes it work well when the network size is small, and the use of individual reward makes it scale well with a larger $N$.
\end{itemize}
We have done additional experiments to understand the different learning rules better. Experiments on $p$-order Weight Maximization for $p \in \{1, 2, ..., 8\}$ can be found in Appendix \ref{sec:a31}. Experiments on variants of unbiased Weight Maximization with a reduced variance can be found in Appendix \ref{sec:a32}. Experiments on comparison with continuous-valued ANNs trained by backprop can be found in Appendix \ref{sec:a33}. Finally, experiments on the estimation bias and variance of the different estimators can be found in Appendix \ref{sec:a34}.

\section{Discussion}

Both the theoretical results and experiment results of unbiased Weight Maximization are promising. Among all learning rules considered in this report, it is the only unbiased learning rule that scales well in term of performance when the network grows larger. To our knowledge, this is also the first learning rule for a network of Bernoulli-logistic units that have such property.

However, there are a few drawbacks of unbiased Weight Maximization. First, the computation cost to compute the individual reward for a layer is $\mathcal{O}(m^2n)$, where $m$ denotes the number of units on that layer and $n$ denotes the number of units on the next layer, since we have to compute the sampling ratio for each unit and each computation of sampling ratio requires $\mathcal{O}(mn)$. In contrast, the computation cost to compute the individual reward signal or the feedback signal for a layer is only $\mathcal{O}(mn)$ in Weight Maximization or STE backprop; and no such computation is required for REINFORCE. Second, the computation of individual reward for unbiased Weight Maximization is too involved for biological neurons - it requires all units in a layer to coordinate together to compute the sampling ratio for individual reward signals to the upstream units. REINFORCE or Weight Maximization does not have such issue. Weight Maximization works well when the outgoing weight's norm is small, and the efficacy of synaptic transmission in biological neurons cannot be unbounded, so Weight Maximization, which can be seen as the first-order approximation of unbiased Weight Maximization, may already be sufficient for biological neurons.

An important and fundamental question in this study is the motivation for discrete-valued ANNs in general. A continuous-valued ANNs trained by backprop can learn faster and has a better asymptotic performance than the discrete-valued ANNs trained by all learning rules considered in this report (see Appendix \ref{sec:a33}), and so the motivation of discrete-valued ANNs remains to be explored. 

In conclusion, this report analyzes different learning rules for discrete-valued ANNs
and proposes two new learning rules, called high-order Weight Maximization and unbiased Weight Maximization, that extend from Weight Maximization. Possible future work includes:
\begin{itemize}
	\item Perform more experiments to understand the advantages or disadvantages of unbiased Weight Maximization (e.g.\ generalization error, dynamics of exploration) on standard supervised learning tasks and reinforcement learning tasks.
	\item Generalize unbiased Weight Maximization to other discrete-valued units besides Bernoulli-logistic units, such as softmax units.
	\item Adjust unbiased Weight Maximization to encourage exploration, e.g.\ adding some terms in the learning rule to encourage exploration such as the use of $\lambda$ in $A_{r-p}$ \cite{barto1985learning}.
	\item Explore methods to reduce the computational cost of unbiased Weight Maximization or simplify unbiased Weight Maximization while maintaining the performance or the unbiased property.
	\item Develop an unbiased version of STE backprop, which is straightforward using the idea of unbiased Weight Maximization since both rely on estimates of $r(1) - r(0)$.
\end{itemize}

\section{Afterword: my view on discrete-valued ANNs}

The dynamic of discrete-valued ANNs is very different than continuous-valued ANNs and may lead to a much more powerful network. For example, a restricted Boltzmann machine (RBM) with continuous-valued hidden and visible units can only model multi-variate Gaussian distribution, but an RBM with binary-valued hidden units (which becomes Bernoulli-logistic units) and continuous-valued visible units can model any distributions. Thus, by restricting the activation values of units to binary values, we can get a more powerful network paradoxically. We also observe this phenomenon in some recent advances in deep learning. The attention module in Transformers \cite{vaswani2017attention}, which led to great success in natural language process (NLP) and underlies virtually all models in NLP nowadays, can be seen as a pseudo-discrete operation: the attention vector from a trained attention module is close to a one-hot vector, and so the network can `attend' to different parts of the input in a discrete manner. Nonetheless, due to the lack of methods to train discrete-valued ANNs efficiently, we can only convert discrete operations to differentiable operations, e.g.\ through outputting expected values instead of sampled values as in the attention modules of Transformers, and train them by backprop. However, this may not be desirable - when being asked to think about the favorite fruit, we would not think about an apple with $0.5$ intensity, an orange with $0.3$ intensity, and a banana with $0.2$ intensity at the same time; we think of just one fruit at any moment! Clearly, we do not learn by making every thought differentiable. From a microscopic perspective, the output of a single biological neuron is also not differentiable as it either fires or not. Therefore, it may be better to find methods to directly train discrete operations embedded in an ANN instead of trying to design differentiable variants of them. Some discussions on the benefits of discrete-valued representation can be found in \cite{ferrone2020symbolic, cartuyvels2021discrete}. To conclude, the assumption of everything being differentiable, which virtually underlies all recent advances in deep learning, should be questioned. 

\section{Acknowledgment}

We would like to thank Andrew G. Barto, who inspired this research and provided valuable insights
and comments.

\clearpage
\appendix

\section{A general formula of $r^{(p)}(h)$} \label{sec:a1}

To obtain a general formula for an unbiased estimator of $r^{(p)}(h)$, we first define $P(k)$ to be the set of partition of integer $k$, with each partition represented by a tuple of the number of appearance. For example, $4$ can be partitioned into $4, 3+1, 2+2, 2+1+1, 1+1+1+1$. To encode $2+1+1$, we see that $1$ appears twice, $2$ appears once, so it is encoded into $(2, 1, 0, 0)$. So $P(4) =\{ \{0, 0, 0, 1\}, \{1, 0, 1, 0\}, \{0, 2, 0, 0\}, \{2, 1, 0, 0\}, \{4, 0, 0, 0\}\}$. Then, we define $t_f^{(k)}$ for $k=1, 2, ..., $ by:
\begin{align}
	t_f^{(k)} &= \sum_{i \in P(k)} \frac{n!}{i_1! i_2! ... i_n!}  \left(\frac{t^{(1)}}{1!}\right)^{i_1} \left(\frac{t^{(2)}}{2!}\right)^{i_2} ... \left(\frac{t^{(k)}}{k!}\right)^{i_k},
\end{align}
and $t_f^{(0)}  = 1$. For example, for $k=1$ to $4$:
\begin{align}
	t_f^{(1)} &= t^{(1)},  \\
	t_f^{(2)} &= t^{(2)} + (t^{(1)})^2,  \\
	t_f^{(3)} &= t^{(3)} + 3 t^{(1)}  t^{(2)} +  (t^{(1)})^3, \\		
	t_f^{(4)} &= t^{(4)} + 4 t^{(1)} t^{(3)} + 3 (t^{(2)})^2 + 6 (t^{(1)})^2 t^{(2)} + (t^{(1)})^4.
\end{align}
And $r^{(p)}(h)$ can be expressed by (can be proved by induction; proof omitted here):
\begin{align}
	r^{(p)}(h) = \ex \left[\sum_{k=1}^p C^{p-1}_{k-1} s^{(k)} t_f^{(p-k)} \bigg| H=h \right], \label{eq:a11}
\end{align}
where $s^{(k)}$ and $t^{(k)}$ are defined in (\ref{eq:hwm1}) and (\ref{eq:hwm2}). Therefore, the general formula to estimate $r^{(p)}(h)$ is $\hat{r}^{(p)}(h) := \sum_{k=1}^p C^{p-1}_{k-1} s^{(k)} t_f^{(p-k)} $. For example, for $k=1$ to $4$ (note that they are the same as the estimates derived previously):
\begin{align}
	\hat{r}^{(1)}(h) &= s^{(1)},  \\
	\hat{r}^{(2)}(h) &= s^{(2)} + t_f^{(1)}, \\
	\hat{r}^{(3)}(h) &= s^{(3)} + 2 s^{(2)} t_f^{(1)} + s^{(1)} t_f^{(2)}, \\
	\hat{r}^{(4)}(h) &= s^{(4)} + 3 s^{(3)} t_f^{(1)} + 3 s^{(2)} t_f^{(2)} + s^{(1)} t_f^{(3)}.		
\end{align}

\section{A general formula of $\sigma^{(p)}(x)$} \label{sec:a2}

The general formula of $\sigma^{(p)}(x)$ is given by:
\begin{align}
	\sigma^{(p)}(x) = \sum_{k=1}^p (-1)^{k-1} A_{p,k-1}	\sigma(x)^k (1-\sigma(x))^{p+1-k},
\end{align}
where $A_{p,k-1}$ are the Eulerian Numbers. For the proof, see \cite{minai1993derivatives}. 

To see that $\frac{C_{p+1}\vert\vert v \vert \vert ^{p+1} }{(p+1)!}$ is unbounded if $\vert\vert v \vert \vert > \pi$, it suffices to prove that for $|x| > \pi$,
\begin{align}
	 	\lim_{n \rightarrow \infty} \left| \frac{\sigma^{(2n-1)}(0) x^{(2n-1)}}{(2n-1)!} \right| = +\infty.
\end{align}
\begin{proof}
First, we can write $\sigma^{(p)}(0)$ as:
\begin{align}
	\sigma^{(p)}(0) &= \sum_{k=1}^p (-1)^{k-1} A_{p,k-1}	\sigma(0)^k (1-\sigma(0))^{p+1-k}, \\
	&= \sum_{k=1}^p (-1)^{k-1} A_{p,k-1} 2^{-(p+1)} \\
	&= 2^{-(p+1)} \sum_{k=1}^p (-1)^{k-1} A_{p,k-1} \\	
	&= \frac{(2^{p+1} - 1) B_{p+1}}{p+1}, 
\end{align}
where $B_{p+1}$ denotes the Bernoulli number. Substituting back, we have
\begin{align}
	\lim_{n \rightarrow \infty} \left|\frac{\sigma^{(2n-1)}(0) x^{(2n-1)}}{(2n-1)!} \right| &= 	\lim_{n \rightarrow \infty}  \left|\frac{(2^{2n} - 1) B_{2n} x^{(2n-1)}}{(2n)! }\right| \\
	&\geq \lim_{n \rightarrow \infty}  \left|\frac{(2^{2n} - 1) 2 (2n)! x^{(2n-1)}}{(2\pi)^{2n} (1-2^{-2n})(2n)! }\right| \label{eq:a22}\\
	&=\lim_{n \rightarrow \infty}  \left|\frac{(2^{2n} - 1) 2  x^{(2n-1)}}{(2\pi)^{2n}  }\right| \\
	&=\lim_{n \rightarrow \infty}  \left|\frac{2}{x} \left(\frac{x}{\pi}\right)^{2n} \left(1 - \frac{1}{2^{2n}}\right)\right|,
\end{align}
which is unbounded if $|x| > \pi$. (\ref{eq:a22}) uses $|B_{2n}| > \frac{2 (2n)!}{(2\pi)^{2n} (1-2^{-2n})}$ \cite{alzer2000sharp}.
\end{proof}

\section{Additional Experiments}
\subsection{High-order Weight Maximization} \label{sec:a31}
\begin{figure}[h!!!]
	\centering
	\includegraphics[width=\textwidth]{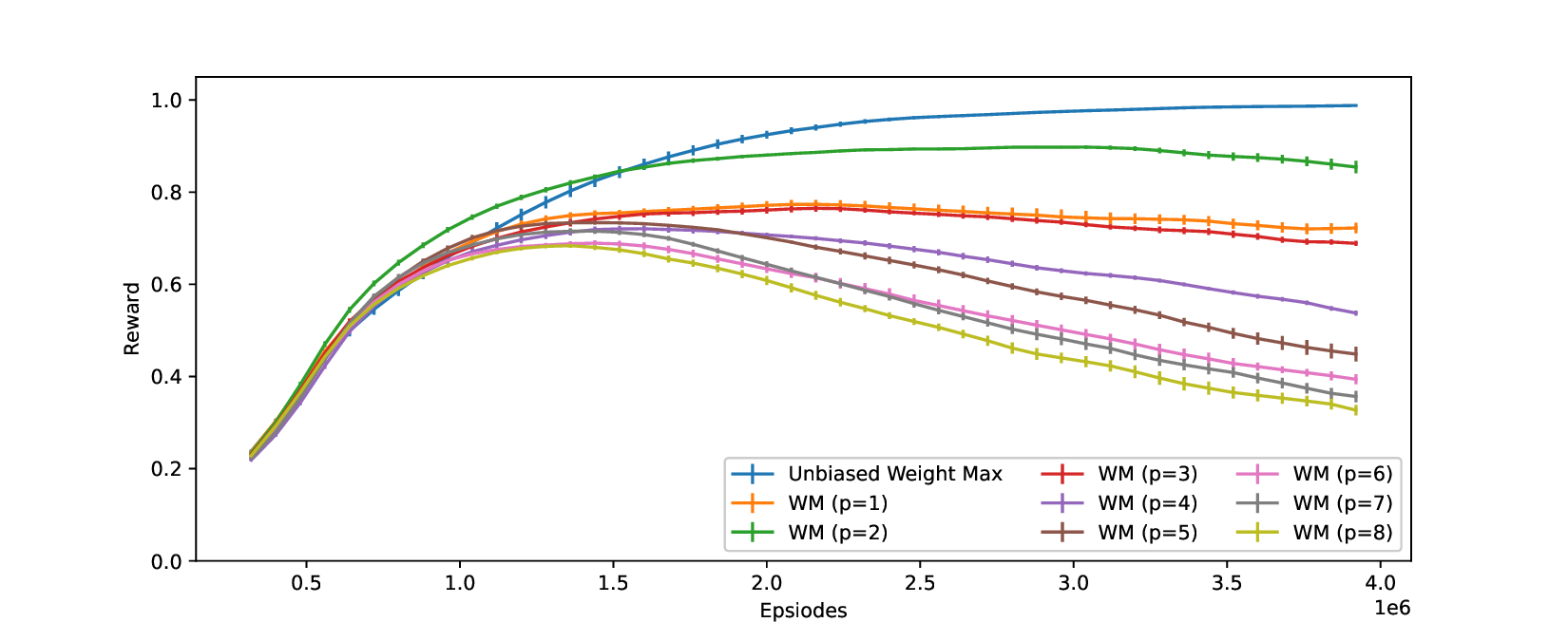}
	\caption{Running average of rewards in each episode throughout a run in the multiplexer task for high-order Weight Maximization. Results are averaged over 5 independent runs, and the error bar represents the standard deviation over the runs.}
	\label{fig:5}
\end{figure}

We repeat the same experiment in Section \ref{sec:ex} for $p$-order Weight Maximization, with $p \in \{1, 2, ..., 8\}$. We use $N=64$. The result is shown in Figure \ref{fig:5}. 

We see that $p$-order Weight Maximization performs worse with a larger $p$. Manual inspection shows that the individual reward computed by high-order Weight Maximization can be very large if the outgoing weight's norm is large. This is essentially due to the problem of approximating the sigmoid function $\sigma(x)$ with Taylor series evaluated $x=a$, which diverges when $|x-a|$ is too large.

\subsection{Variants of unbiased Weight Maximization} \label{sec:a32}

\begin{figure}[h!!!]
	\centering
	\includegraphics[width=\textwidth]{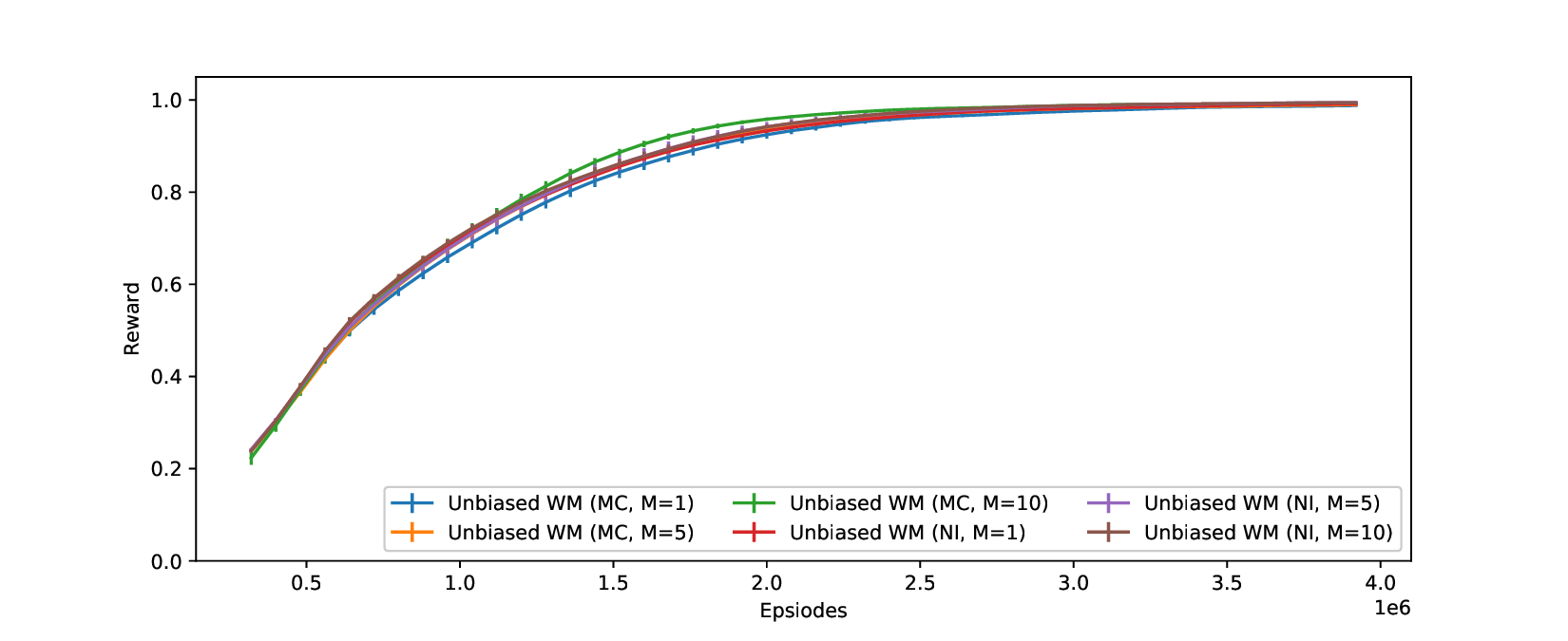}
	\caption{Running average of rewards in each episode throughout a run in the multiplexer task for variants of unbiased Weight Maximization. Results are averaged over 5 independent runs, and the error bar represents the standard deviation over the runs.}
	\label{fig:6}
\end{figure}

In unbiased Weight Maximization we estimate $\ex [r'(U)]$ with $r'(U)$, which is a Monte Carlo estimate with a single sample. An estimate with a lower variance is $\sum_{i=1}^M \frac{r'(U_i)}{M}$, where all $U_i$ are i.i.d. and uniformly distributed. Alternatives, we can estimate $\ex [r'(U)]$ by numerical integration using the rectangle rule applied on $M$ subintervals. To test these methods, we repeat the same experiment in Section \ref{sec:ex} with $M \in \{1, 5, 10\}$ for both Monte Carlo (MC) and numerical integration (NI). We use $N=64$. The result is shown in Figure \ref{fig:6}. 

We see that the learning curves are almost the same. It seems that these different methods to reduce variance do not yield observable improvement in performance despite the reduced variance. Maybe the reduction in variance is not significant enough to affect performance.

\subsection{Comparison with continuous-valued ANNs} \label{sec:a33}

\begin{figure}[h!!!]
	\centering
	\includegraphics[width=\textwidth]{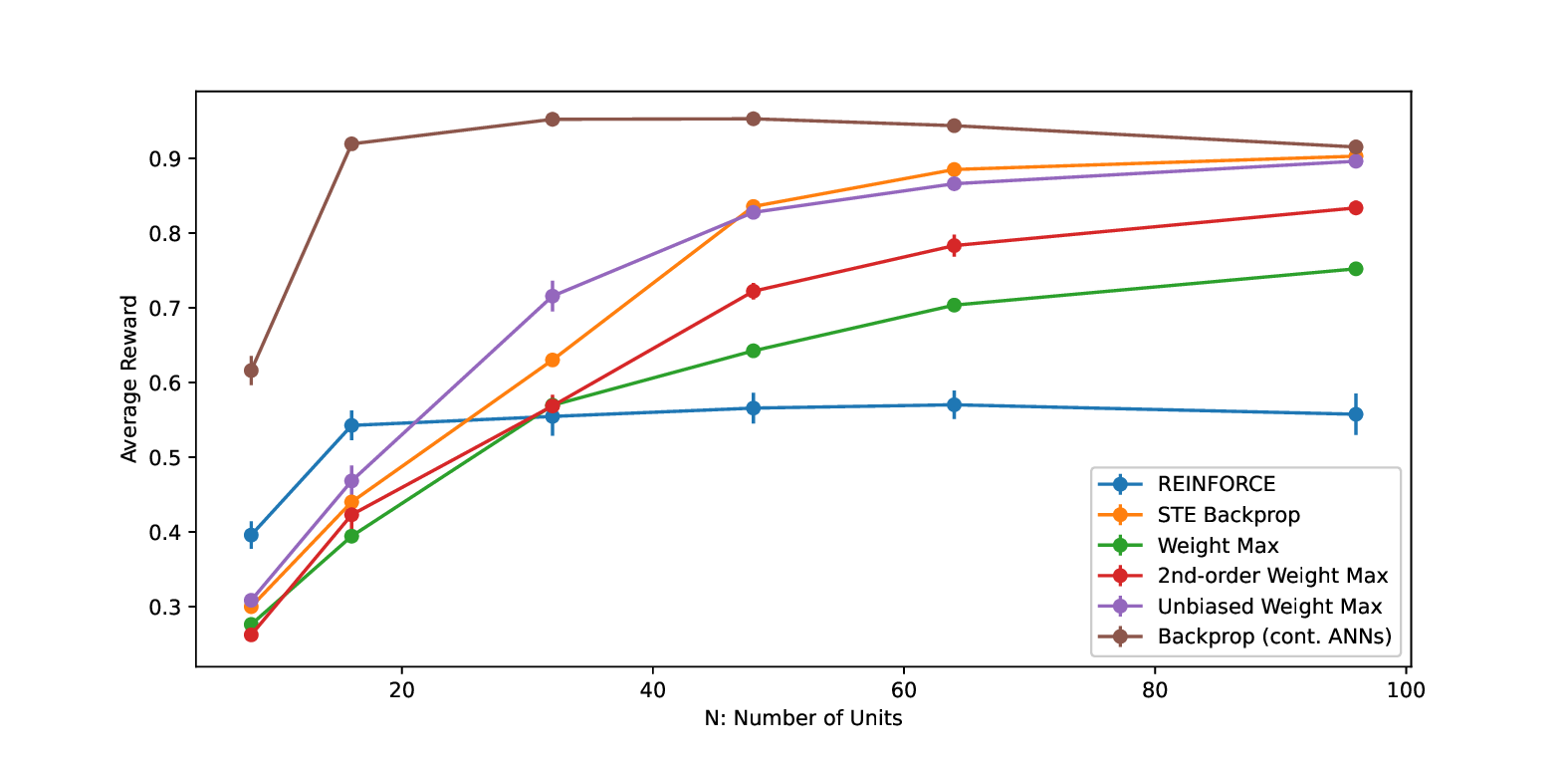}
	\caption{Average rewards of the first $5e6$ episodes in the multiplexer task with a different $N$ for both discrete-valued and continuous-valued ANNs. Results are averaged over 5 independent runs, and the error bar represents the standard deviation over the runs.}
	\label{fig:7}
\end{figure}

\begin{figure}[h!!!]
	\centering
	\includegraphics[width=\textwidth]{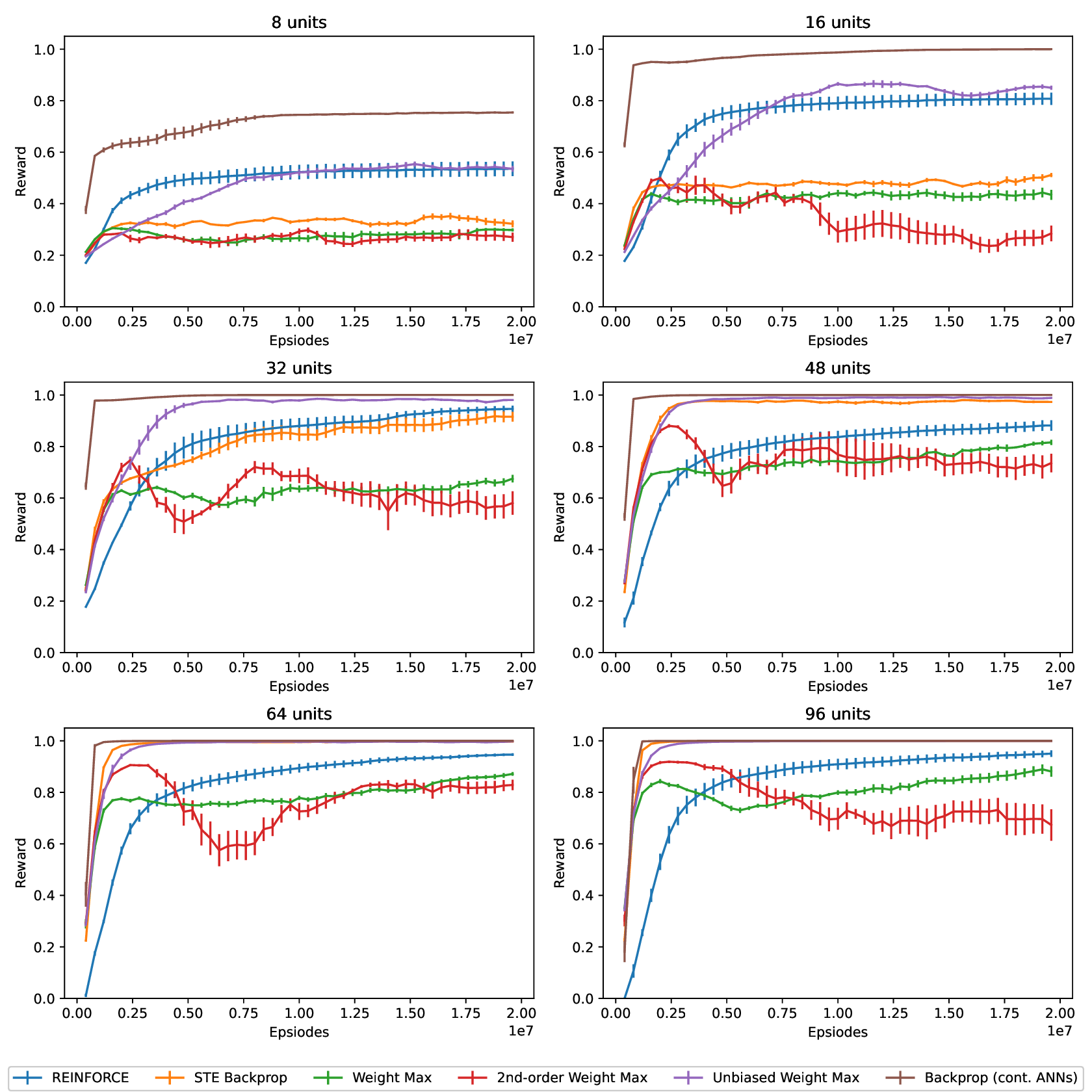}
	\caption{Running average of rewards in each episode throughout a run in the multiplexer task with a different $N$ for both discrete-valued and continuous-valued ANNs. Results are averaged over 5 independent runs, and the error bar represents the standard deviation over the runs.}
	\label{fig:8}
\end{figure}

We repeat the same experiment in Section \ref{sec:ex} for continuous-valued ANNs trained by backprop: the output unit is still a Bernoulli-logistic unit trained by REINFORCE, but all hidden units are deterministic sigmoid units trained by backprop. The architecture of the network is the same as the discrete-valued ANNs. Thus, the continuous-valued ANN considered is equivalent to the discrete-valued ANN in our experiments but with hidden units outputting expected values instead of sampled values. The average reward, similar to Figure \ref{fig:3} and Figure \ref{fig:4}, is shown on Figure \ref{fig:7} and Figure \ref{fig:8}.

We observe that continuous-valued ANNs trained by backprop has a better performance in term of both learning speed and asymptotic performance than discrete-valued ANNs, though the difference narrows for a larger network. This is likely because units in discrete-valued ANNs can only communicate by binary values, which constraints the amount of information passed between units, e.g.\ a unit can only express if there is an edge but not the probability that there is an edge. As the network size increases, more binary units can detect the same edge, and the average value of these units can estimate the probability that there is an edge, so the difference in performance narrows.

More experiments can be conducted to evaluate and compare discrete-valued ANNs and continuous-valued ANNs on other aspects besides learning speed and network capacity, such as generalization error in supervised learning tasks and dynamics of exploration in reinforcement learning tasks.

\subsection{Estimation bias and variance of the estimators} \label{sec:a34}

All learning rules discussed can be seen as using different estimators to estimate the gradient of expected reward $\ex[r(H)]$ w.r.t. bias of the unit $b$. To evaluate the quality of estimators, we compute their estimation bias and variance on some randomly selected distributions. 

We consider a five-layered network of Bernoulli-logistic units. The first hidden layer and the output layer have a single unit; the other hidden layers have four units. The weight and bias of these units are all uniformly sampled from $[-C, +C]$ where $C > 0$. The task considered has no inputs, and the rewards for the two binary outputs are uniformly sampled from $[-10, +10]$. Then we apply the estimators to estimate $\ex[r(H)]$ w.r.t. bias of the unit on the first layer. The estimation bias and variance of the different estimators can be analytically computed as the network's size is small. 

\begin{figure}[h!!!]
	\centering
	\includegraphics[width=\textwidth]{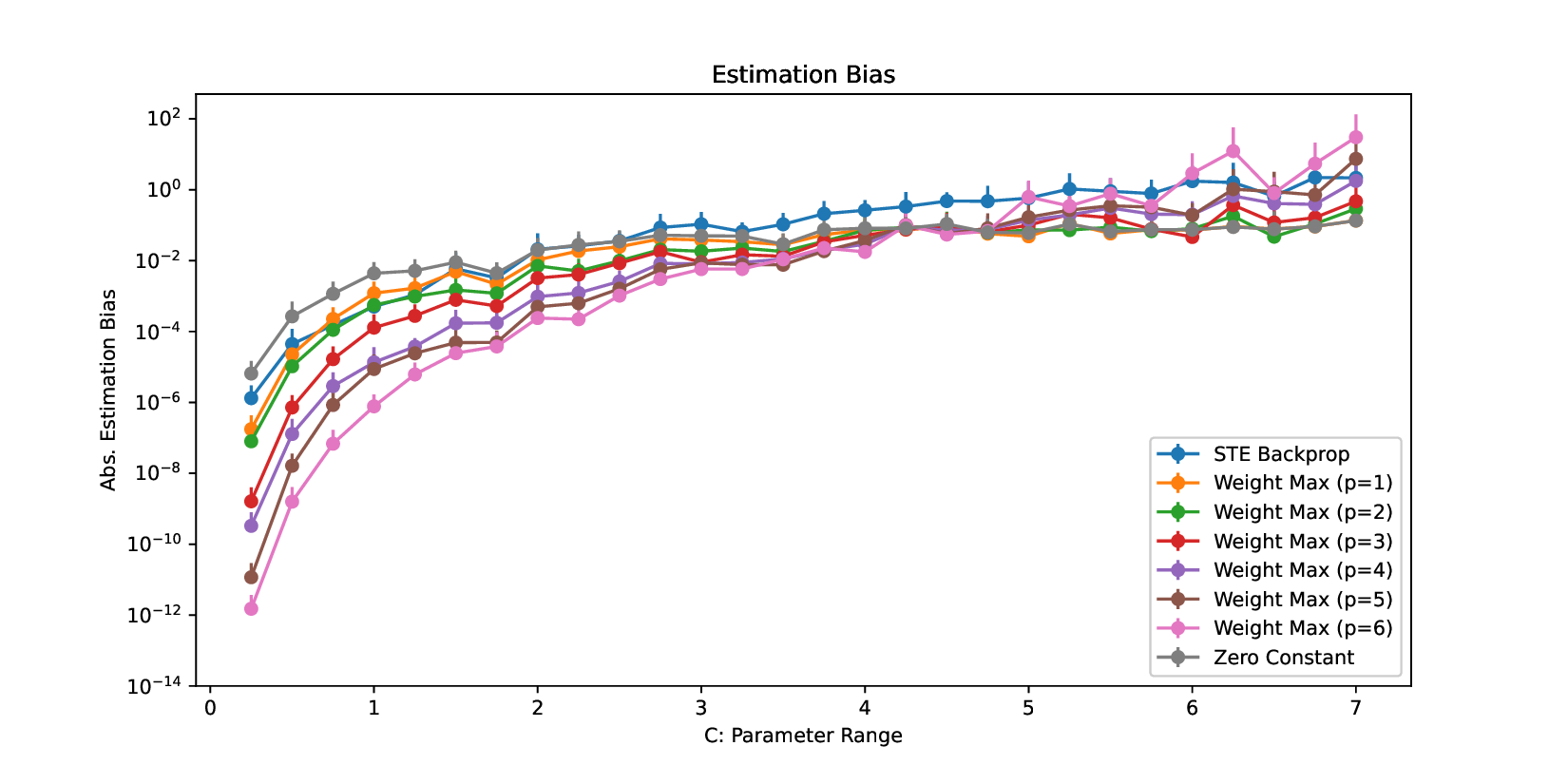}
	\caption{Estimation bias of different estimators in a network with different parameter range $M$. Note that REINFORCE and unbiased Weight Maximization have zero estimation bias, and therefore are not shown on the graph. Results shown on the graph are averaged over 20 randomly selected distribution.}
	\label{fig:9}
\end{figure}

\begin{figure}[h!!!]
	\centering
	\includegraphics[width=\textwidth]{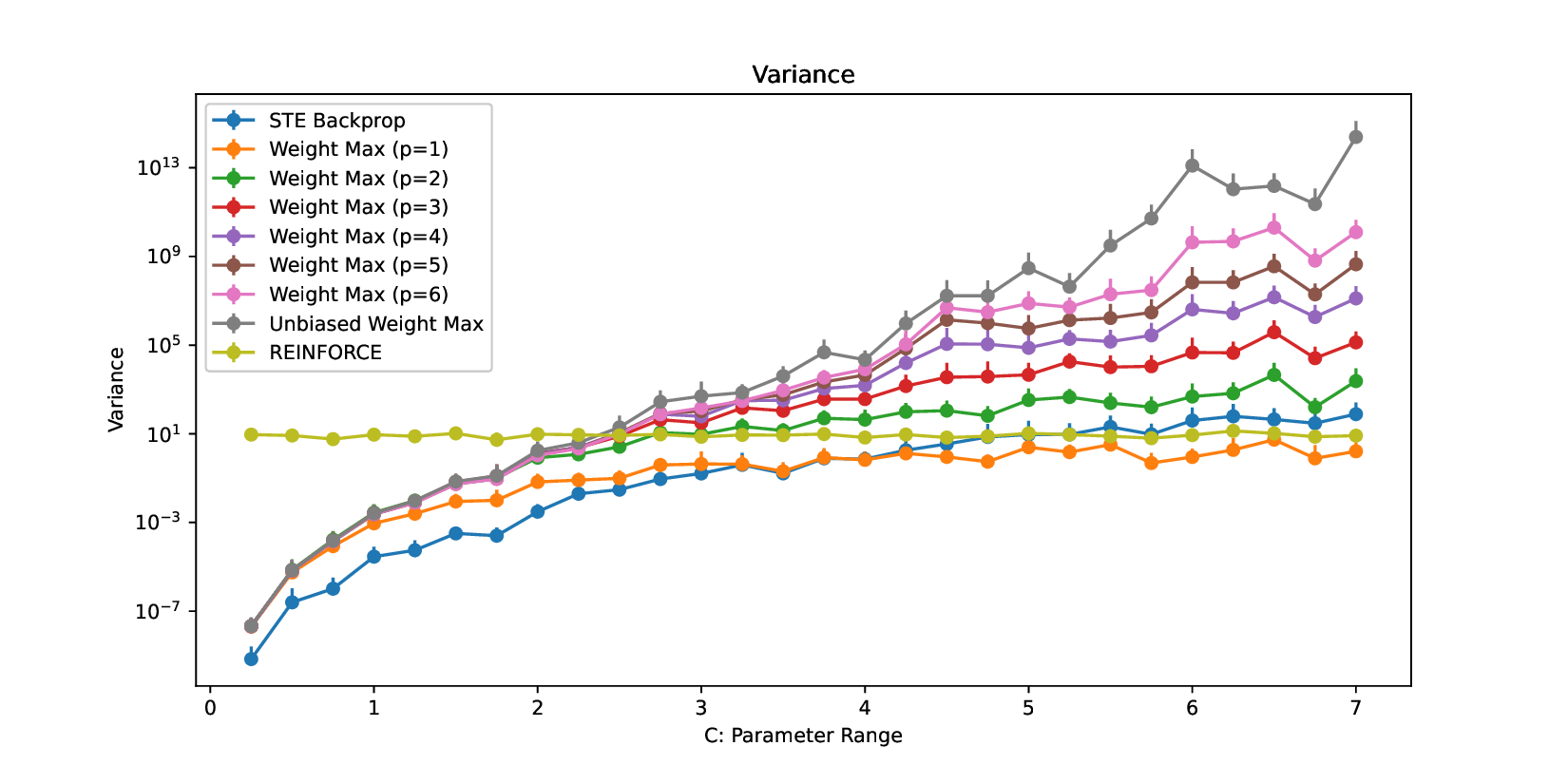}
	\caption{Variance of different estimators in a network with different parameter range $M$. Results shown on the graph are averaged over 20 randomly selected distribution.}
	\label{fig:10}
\end{figure}

The estimation bias and variance for a network with different parameter range $C$ are shown in Figure \ref{fig:9} and Figure \ref{fig:10}. We observe that when the $C$ is small, the estimation bias of $p$-order Weight Maximization is lower for a higher $p$ - at that range, the Taylor series of sigmoid function converges. However, when $C$ is large, the Taylor series diverges, and so the estimation bias is larger for a higher $p$. As for variance, we observe that REINFORCE has a steady variance regardless of $C$, since the variance of REINFORCE is bounded by the square of expected rewards. Unbiased Weight Maximization's variance increases fastest with $C$, since the sampling ratio in unbiased Weight Maximization can be huge when the weight is large. Nonetheless, the large variance of unbiased Weight Maximization seems to have not much impact on the learning curve. This may be explained by the fact that those huge sampling ratios only occur with a probability close to zero and so may never be encountered during training. The relationship between variance and learning speed deserves further investigation.

\clearpage
\printbibliography
\end{document}